\documentclass[12pt]{article}
\usepackage{caption}
\usepackage{subcaption}
\usepackage[dvipdfm]{graphicx} 
\usepackage{bmpsize}
\usepackage{pdfpages}
\usepackage{amsmath}
\usepackage{times}
\usepackage{graphicx}
\usepackage{color}
\usepackage{rotating}
\usepackage{bbm}
\usepackage{latexsym}
\usepackage{epsfig}
\usepackage{amssymb}
\usepackage{slashbox}

\usepackage[noend]{algorithmic} 
\usepackage{algorithm,caption}
\algsetup{indent=2em} 
 
\algsetup{indent=2em} 
\begin{document}

\textbf{Variational Mixture Models with Gamma or inverse-Gamma components}\\
\begin{center} A. Llera$^{1}$, D. Vidaurre$^{2}$, R.H.R. Pruim$^{1}$, C. F. Beckmann$^{1}$. \end{center}
\begin{center} {\footnotesize Technical report \\ 1- Donders Institute for Brain Cognition and Behaviour, Radboud University Nijmegen \\
2- Oxford centre for Human Brain Activity (OHBA)}\end{center} 

\begin{center} {\bf Abstract} \end{center}
Mixture models with Gamma and/or inverse-Gamma distributed mixture components are useful for medical image tissue segmentation or as post-hoc models for regression coefficients obtained from linear regression within a Generalised Linear Modeling framework (GLM), used in this case to separate stochastic (Gaussian) noise from some kind of positive or negative 'activation' (modeled as Gamma or inverse-Gamma distributed). 
To date, the most common choice in this context it is Gaussian/Gamma mixture models learned through a maximum likelihood (ML) approach; we recently extended such algorithm for mixture models with inverse-Gamma components. Here, we introduce a fully analytical Variational Bayes (VB) learning framework for both Gamma and/or inverse-Gamma components.

We use synthetic and resting state fMRI data to compare the performance of the ML and VB algorithms in terms of area under the curve and computational cost. 
We observed that the ML Gaussian/Gamma model is very expensive specially when considering high resolution images; furthermore, these solutions are highly variable and they occasionally can overestimate the activations severely. The Bayesian Gauss/Gamma is in general the fastest algorithm but provides too dense solutions. The maximum likelihood Gaussian/inverse-Gamma is also very fast but provides in general very sparse solutions. The variational Gaussian/inverse-Gamma mixture model is the most robust and its cost is acceptable even for high resolution images. 
Further, the presented methodology represents an essential building block that can be directly used in more complex inference tasks, specially designed to analyse MRI/fMRI data; such models include for example analytical variational mixture models with adaptive spatial regularization or better source models for new spatial blind source separation approaches.

\section{Introduction}
Mixture models are an important and powerful tool in many practical applications thanks to their ability to flexibly model complex data \cite{lindsay95}. Mixture models containing Gamma or inverse-Gamma distributed components are interesting due to the positive support of such distributions and are commonly used to provide class-dependent models separating stochastic noise, typically modeled by a close to zero-mean Gaussian distribution, from some kind of activation modeled by a positive support distribution \cite{Lit:Gm-N-Gm_1,Lit:invg1,Lit:invgm2}. For example, in medical imaging such models can be used for statistical segmentation of structural images into different tissue types on the basis of measured intensity levels. Also, in functional statistical parametric mapping (where voxels are either activated or not activated), mixture models can be used for post-hoc inference on the regression maps\cite{advancesFSL}. 

The most common approach to learn mixture models in general is the expectation maximization EM algorithm (EM) which is used to estimate a maximum likelihood (ML) solution \cite{bishop}. However, since there is no closed form ML solution neither for the scale parameter of the Gamma nor for the shape parameter of the inverse-Gamma, the problem becomes more complex and typically requires numerical optimization \cite{Gm-N-Gm_1,gmmm12,Lit:invg1,invg2}. Numerical optimization must be performed at each iteration of the EM algorithm, making such strategy computationally hard, specially for cases where the number of samples is very high, as e.g. high resolution whole brain MRI data. A common faster alternative uses the method of moment approximation to estimate the parameters of the Gamma or inverse-Gamma components \cite{poster,posterohbm2015,fsl}. We denote the algorithm presented in \cite{poster} for learning a Gauss/Gamma mixture model as GGM, and the one presented in \cite{posterohbm2015} for learning  Gaussian/inverse-Gamma ones as GIM. An alternative to such ML approaches is to consider Bayesian inference. The Bayesian approach provides an elegant way to explore uncertainty in the model and/or to include prior knowledge into the learning process. Furthermore, it provides principled model selection to select the number of components in the mixture model, and it allows to use the learnt components as building blocks of bigger Bayesian inference problems \cite{choud}. To the extent of our knowledge, there are sampling algorithms available for the Gamma case \cite{marin,GGM_MCMC} and versions providing spatial regularization \cite{woolMM}. However, the sampling strategy can be computationally infeasible for high resolution images, and specially in cases where the mixture distributions become part of bigger statistical learning problems \cite{Makni_2006a.Bayesian}. Variational Bayes (VB) inference \cite{bishop} provides instead a more efficient alternative.
Althought in \cite{WoolSPMM} a variational Gaussian/Gamma mixture model with spatial regularization is presented, the Gamma distribution parameters of the mixture model are learnt using a conjugate gradient numerical optimization procedure. %No fully variational Gauss/Gamma neither Gauss/inverse-Gamma mixture model have been previously reported in the literature.
In this work we introduce novel algorithms for learning mixture models with Gamma and/or inverse-Gamma components using an analytic VB approach. While most parameters belong to conjugate distributions and can be estimated easily, learning the shape parameter of the distributions is not so straightforward. For the shape parameters we use unnormalized conjugate priors \cite{Fink97acompendium,LleBe_arxiv_2016_1}, resorting to Laplace approximations and Taylor expansions to compute the required posterior expectations.

%While for Gamma components there is an unnormalized conjugate prior for the shape parameter \cite{Fink97acompendium}, there exists no known conjugate prior for the inverse-Gamma shape. Here we introduce a novel unnormalized conjugate prior for the inverse-Gamma shape parameter and for both cases, Gamma or Inverse-Gamma components, we use the Laplace approximation and Taylor expansions to compute the necessary posterior expectations over the shape parameters. 

In section \ref{methods_sec}, we introduce the four considered models and outline the datasets used to evaluate them. 
In section \ref{model_sec}, we introduce the basic notation and a brief description of the learning algorithms. Further details are given in the Appendix. 
In section \ref{Syn0}, we describe the synthetic data sets we consider for evaluation of the methods. 
In section \ref{rfMRI} we describe the resting state fMRI dataset as well as the data processing performed to obtain 4400 spatial maps extracted from 100 subjects rfMRI data. 
In sections \ref{numres1} and \ref{val_rfMRI} we present the results obtained by comparing the two newly proposed algorithms with their maximum likelihood counterparts in both artificial and rfMRI data. 
Finally, in section \ref{Discussion}, we conclude the paper with a brief discussion.

\section{Methods}
\label{methods_sec}
We now introduce the methodology and the datasets used to evaluate the different considered models. In section \ref{model_sec}, we introduced the notation necessary to describe the problem and, in section \ref{model_sec2}, we introduced the two state of the art models alongside their two new Bayesian versions. Then, we introduce the (synthetic and rfMRI) datasets that will later be used to evaluate the four considered models. 

\subsection{The problem}
\label{model_sec}
\begin{figure}
\centering
\begin{subfigure}{.5\textwidth}
  \centering
  \includegraphics[width=.8\linewidth]{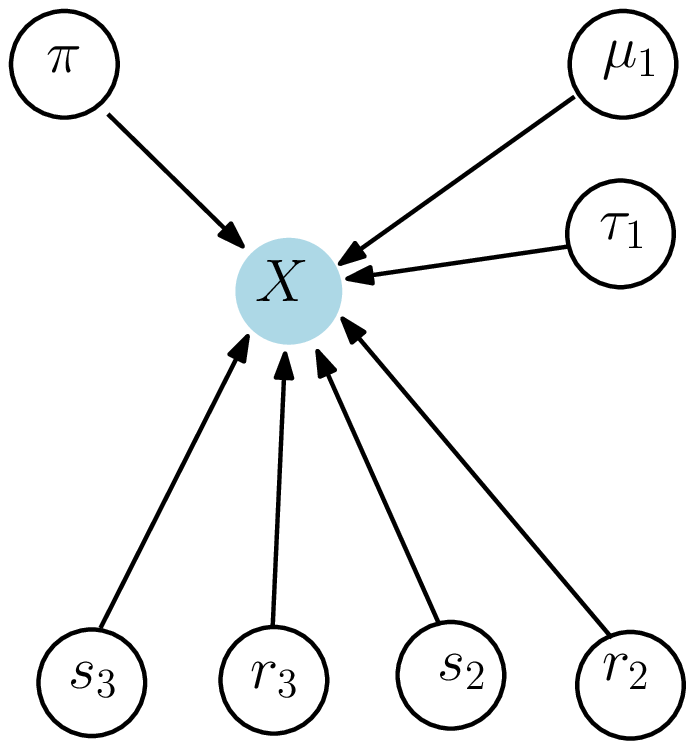}
  %\caption{A subfigure}
  %\label{fig1:sub1}
\end{subfigure}%
\begin{subfigure}{.5\textwidth}
  \centering
  \includegraphics[width=.8\linewidth]{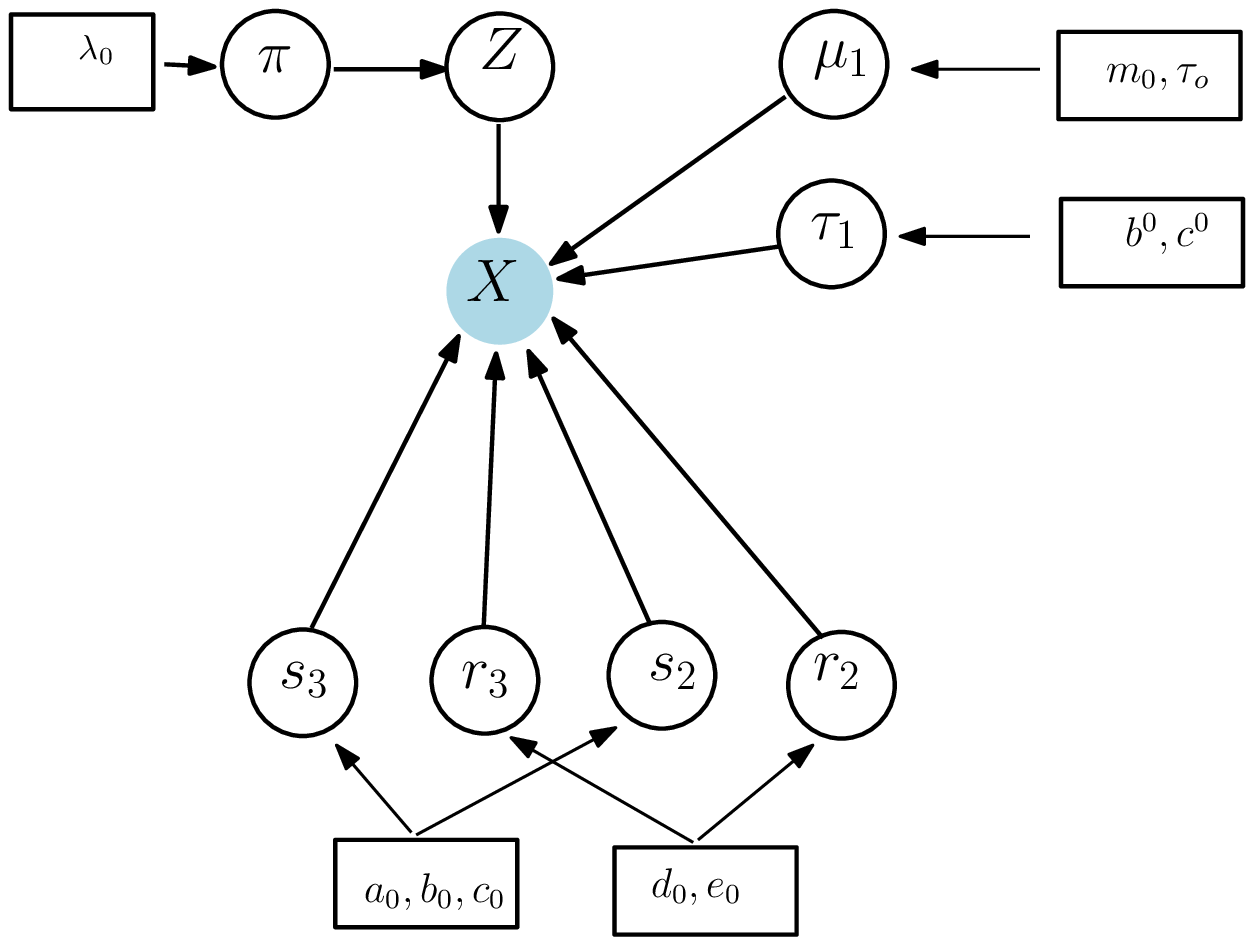}
  %\caption{A subfigure}
  %\label{fig1:sub2}
\end{subfigure}
\caption{Figure 1 left panel shows a graphical representation of a mixture model with 3 components, one Gaussian and two Gamma and/or inverse-Gamma distributed. The right panel shows such a representation when including prior distributions over the mixture model parameters as well as an indicator variable Z (see text for more details).}
\label{fig1}
\end{figure}

Let $\boldsymbol{x}=\{x_1,\ldots,x_N \}$, $x_i \in \mathbb{R}$ be an observation data vector. Without algorithmic loss of generality we will reduce derivations to mixture models of three components, so that $p(\boldsymbol{x}|\boldsymbol{\pi}, \Theta)=\prod_{n=1}^{N} \sum_{k=1}^{3} \pi_k p_k(x_n|\Theta_k)$, where $\Theta=\{\Theta_1, \Theta_2, \Theta_3 \}$ are the parameters of the three components and $\boldsymbol{\pi}= \{\pi_1, \pi_2, \pi_3 \}$ are the mixing proportions. One component is used to model stochastic noise which, as usual, is modeled using a Gaussian component: $p_1(x|\mu_1,\tau_1)= \mathcal{N}(x|\mu_1,\tau_1)$ with $\mu \approx 0$. The other two components model independently positive and negative activations. Here we extend the common choice of Gamma distributions to consider also inverse-Gamma components, that means that the  
positive component $p_2(x|s_2,r_2)$ can be chosen to be Gamma 
\begin{align*}
p_2(x|s_2,r_2)= \mathcal{G}(x|s_2,r_2)
\end{align*} 
or inverse-Gamma distributed
\begin{align*}
p_2(x|s_2,r_2)= \mathcal{IG}(x|s_2,r_2),
\end{align*} 
and the negative component $p_3(x|s_3,r_3)$ can be negative Gamma
\begin{align*}
p_3(x|s_3,r_3)= \mathcal{G}^{-}(x|s_3,r_3) = \mathcal{G}(-x|s_3,r_3)
\end{align*} 
or Negative inverse-Gamma distributed
\begin{align*}
p_3(x|s_3,r_3)= \mathcal{IG}^{-}(x|s_3,r_3) = \mathcal{IG}(-x|s_3,r_3).
\end{align*} 
Regardless of the choice of the distribution, $s_k$ represents the shape of the distribution; for any Gamma component $r_k$ denotes the rate parameter, and for the inverse-Gamma ones it denotes the scale parameter. A general graphical representation is presented in the left panel of Figure \ref{fig1}.

\subsection{The solutions}
\label{model_sec2}
Learning the model parameters $\Theta=\{ \boldsymbol{\pi}, \mu_1,\tau_1,\boldsymbol{s}, \boldsymbol{r}\}$ is usually achieved through the EM algorithms presented in \cite{poster,posterohbm2015,fsl}. 
These algorithms use the method of moment approximation for the Gamma or inverse-Gamma component parameters in order to compute the so-called responsibilities and update the expected means and variances analogous to an EM for Gaussian Mixture model \cite{poster,posterohbm2015,fsl}. An alternative to maximum likelihood (ML) approaches is to perform Bayesian inference. Defining prior distributions over each parameter,
the right panel of Figure \ref{fig1} shows a graphical representation for such mixture models where the hyper-priors parameters are represented inside the rectangles. 
We use a Dirichlet prior for the mixing proportions $\boldsymbol{\pi}$, a Gaussian prior for the Gaussian mean $\mu_1$ and a Gamma prior for its precision $\tau_1$. For the $\boldsymbol{r}$ parameters we use a Gamma prior. For the shape parameter of the Gamma we use the unnormalized conjugate prior proposed in \cite{Fink97acompendium} and for the inverse-Gamma the prior we recently introduced in \cite{LleBe_arxiv_2016_1}. 
Note that we also introduced an indicator function $Z$, so that, for each observation $x_n$, we define a latent variable $\mathbf{z}_n$ as a binary vector with elements $z_{nk}$, $k \in \{1,2,3 \}$, such that $\sum_{k=1}^3 z_{nk} = 1$ and we define $\boldsymbol{Z}=\{\mathbf{z}_1, \ldots, \mathbf{z}_N \}$. 
Consequently, the vector $\boldsymbol{z_n}$ has a value of one in the component number to which $x_n$ belongs.% for a given set of parameters.
For a given set of initialization parameters and hyper-parameters values, the posterior expectation on the parameters can in most cases be easily computed by evaluating expectations over well-known distributions. However, computing the shapes posterior expectations is not straightforward. Here, we use Laplace approximations and Taylor expansions to approximate the solution.
In Appendix A we introduce for the first time the detailed methodology that allows us to perform VB inference in such models. We will further denote these algorithms as Algorithm 1 or bGGM and Algorithm 2 or bGIM for the Gamma and inverse-Gamma cases, respectively.
For ease of notation, we will denote the ML Gaussian/Gamma algorithm presented in \cite{poster} as Algorithm 3 or GGM, and the ML Gaussian/inverse-Gamma algorithm presented in \cite{posterohbm2015} as Algorithm 4 or GIM. For completeness, the method of moments identities as well as both ML algorithms are detailed in the Appendices B and C respectively.

\subsection{Synthetic data}
\label{Syn0}
Synthetic dataset I is generated from Gaussian mixture models with three components and different parameter values. One component has always mean zero while the other two have means $SNR$ and -$SNR$ respectively, with $SNR \in \{2,3,4,5\}$. The variance of all components is always one and we consider three different levels of symmetric activation, $\pi \in \{ [.8,.1, .1 ], [.9, .05, .05 ],[.99,.005, .005 ] \}$. These three levels of activation will be denoted respectively as sparsity 1, 2 and 3. In general, a stronger activation makes easier the problem; the range of considered activations was chosen to illustrate a range of problems, from easy at 20 $\%$ activation to difficult at $1\%$. The intermediate proportion (10$\%$) is intended to emulate a strong rfMRI activation.   
At each simulation we generate N=10000 samples (voxels).

We also consider another synthetic dataset, Synthetic dataset II, which is generated similarly to Synthetic dataset I but with mixing proportions \newline $\pi \in \{ [.9,.1, 0 ], [.95, .05, 0 ],[.99,.01, 0 ] \}$. Thus, Synthetic dataset II contains postive activation but no negative activation. 

For each of the synthetic datasets and for each possible of the 12 possible combinations of SNR and mixture proportions, we generated $N$ samples from such mixture model and we repeated the process 100 times.
In all scenarios we fitted mixture models with three components; therefore, Synthetic dataset II is intended to study the performance of the models with a wrong model order. 

\subsection{Resting State fMRI data}
\label{rfMRI}
We use resting state fMRI (rfMRI) data from 100 healthy controls from the NeuroIMAGE project; this subset of healthy subjects has been previously used in \cite{Aroma}. For specific information on the scanning protocol and parameters of the NeuroIMAGE datasets we refer the reader to \cite{neuroimage}. All rfMRI data processing was carried out using tools from the FMRIB Software Library (FSL\footnote{http://www.fmrib.ox.ac.uk/fsl}) \cite{advancesFSL,fsl,fsl2}. The preprocessing involved removal of the first five volumes to allow for signal equilibration, head movement correction by volume-realignment to the middle volume using MCFLIRT \cite{aff_reg}, global 4D mean intensity normalization, 6mm full-width-half-maximum (FWHM) spatial smoothing, ICA-AROMA based automatic removal of residual motion artifacts \cite{Aroma}, nuisance regression (using mean white matter, CSF time-courses and linear trend as nuisance regressors) and temporal high-pass filtering ($\textgreater 0.01$ Hz). 
For each participant we transformed the rfMRI data to his/her structural image using FLIRT \cite{aff_reg}, an affine boundary-based registration. Then, we registered the functional data to the 4mm isotropic resolution MNI152 standard space using a non-linear registration procedure (FNIRT \cite{non-lin-reg}).

To delineate a set of group-level spatial components we conducted a temporal concatenated group-ICA on the preprocessed data using MELODIC \cite{pica}, where the model order was automatically estimated, resulting in a number of 11 components. Individual spatial maps were derived from the group maps using dual regression \cite{dual_reg} for a total of $11 \times 100 = 1100$ spatial maps. 

To compare the performance of the models under different image resolutions we also resampled all these spatial maps to 3mm, 2mm and 1mm isotropic resolution MNI152 standard space, using FLIRT \cite{aff_reg}. Altogether, we have a total of 4400 spatial maps.

\section{Numerical Results}
\label{numres}
In this section we compare the four considered models, bGGM, bGIM, GGM and GIM. In section \ref{numres1}, we evaluate the models using the synthetic data reported in section \ref{Syn0}. In section \ref{val_rfMRI}, we test them on the statistical maps extracted from resting state fMRI as described in section \ref{rfMRI}. In the remaining, we will denote the different models, bGGM, bGIM, GGM and GIM as algorithms 1-4, using the following color code to identify the models: red=bGGM, green=bGIM, pink=GGM and blue=GIM.

\subsection{Results on synthetic data}
\label{numres1}
\begin{figure}[!h]
\centering
\includegraphics[width=.8\textwidth]{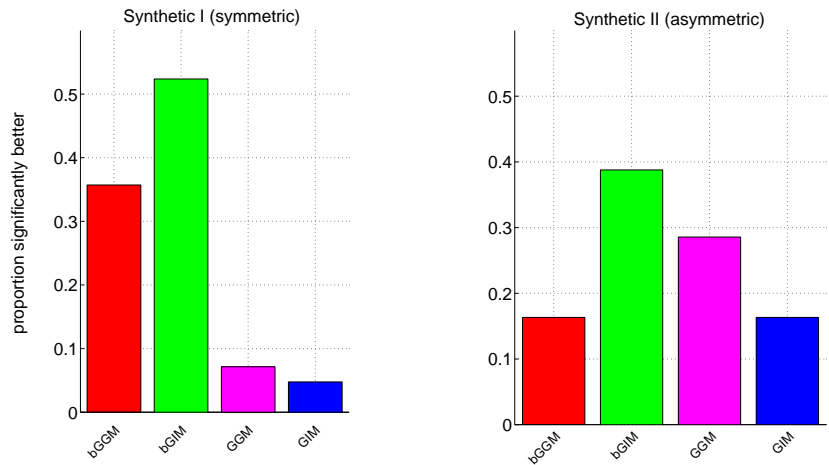}.
\caption{\footnotesize Histogram reflecting the percentage of times each model provides significantly higher normalized AUC than another one. Left pannel shows results in Synthetic dataset 1 (symmetric activation) and the right pannel in synthetic dataset 2 (only positive activation).} 
\label{fig:hist}
\end{figure}

The four considered models are evaluated first in terms of the area under the curve (AUC), normalized in the range FPR $\in [0, 0.05 ]$. In all cases we fitted mixture models with three components. As expected, we observed that all algorithms benefit from a higher SNR and show higher variance at sparser activations (not shown). For each different SNR and mixture proportions, we then compare each pair of models using a paired t-test.
In Figure \ref{fig:hist} we present histograms reflecting the percentage of times a model was significantly better than any other model (statistical significance is considered for p-values $<0.01$). 

The left pannel of figure \ref{fig:hist} presents the results obtained on synthetic dataset I (symmetric activation) and the right one on synthetic dataset II (only positive activation). In the case of synthetic dataset I, we observed that VBGGM and VBGIM were the best models. Further, VBGGM was better than VBGIM at the lowest SNR with strongest activations while VBGIM was better in all the other scenarios. With respect to synthetic dataset II, VBGIM and MLGGM were the best two models and, again, VBGIM was best in most cases with the exception of the low SNR and strong activation cases. 

\begin{figure}[!h]
\centering
\includegraphics[width=1\textwidth]{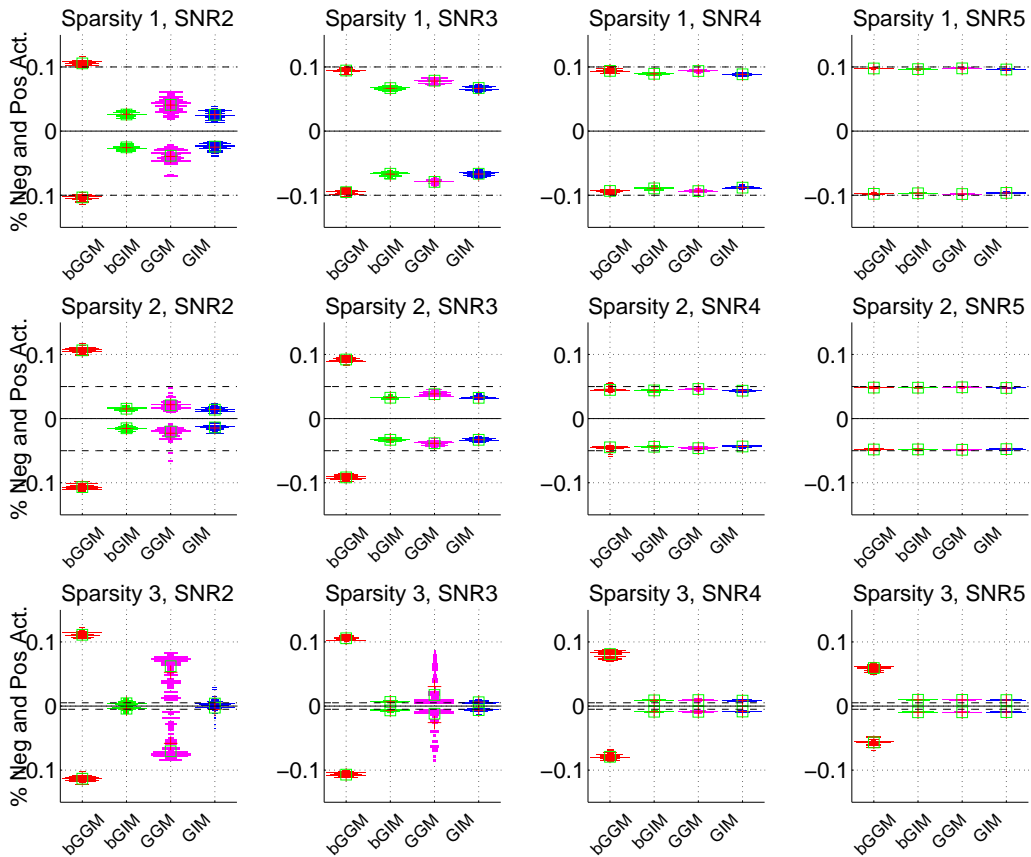}.
\caption{\footnotesize Violin plots of the percentage of positive and negative active voxels of each algorithm (color coded) for Synthetic data I. Each row presents a different symmetric activation levels (or datasets). Each column considers a different SNR. The black discontinuous horizontal line represents the true value. } 
\label{toy1_act1}
\end{figure}

To get a more intuitive idea of the solutions delivered by each model, in Figure \ref{toy1_act1} we present 
violin plots of the percentage of positive and negative active voxels provided by each model when considering synthetic dataset I; for visualization the negative proportion is presented as a negative number. The black discontinuous horizontal lines represent the true activation percentages.
Most models provide generally accurate mixing proportions at high SNR. While GIM provides generally very sparse solutions, GGM shows the highest variance in the solutions and overestimates activations specially at low SNR (first column) or sparse activations (last row). 
Relating the Bayesian models, the bottom row shows that in scenarios where activation is very sparse, the variational Gamma model, bGGM, overestimates activations even at high SNR; the bGIM solution is sparser than the Gamma models and it is very robust as reflected by the low variance in the solutions shown at all SNR and different mixture proportions. 
Although AUC indicates that bGGM is often an appropriate model, this algorithm overestimates activations at sparser cases. This seemingly contradictory effect occurs because, although the Gamma distribution might overestimate the activation, it still models fairly well the tail of the distribution, which is reflected in the restricted AUC measure. Note that the restricted AUC is a reasonable validation measure when considering fMRI data since more than $5\%$ of false positives would provide meaningless results.   

\begin{figure}[!h]
\centering
\includegraphics[width=1\textwidth]{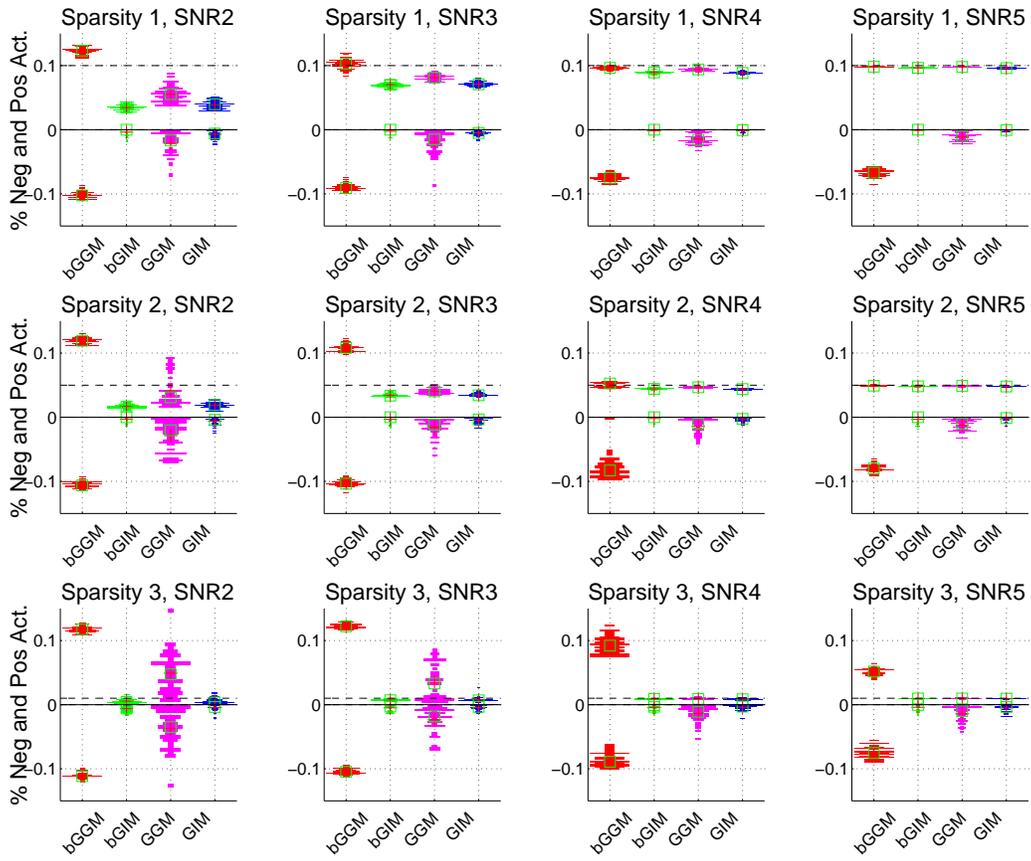}.
\caption{\footnotesize Violin plots of the AUC of each algorithm (color coded) evaluating synthetic data II. Each row corresponds to a different level of sparsity and each column corresponds to a different SNR.} 
\label{fig:un}
\end{figure}
In Figure \ref{fig:un} we present  violin plots of the percentage of positive and negative active voxels when considering synthetic dataset II. As before, for ease of visualization, the negative proportion is presented as a negative number in every different scenario. The black discontinuous horizontal line marks the true activations percentage at each dataset. Note that the only difference between this dataset and the previous synthetic data I is that synthetic data II contains no negative activation. Thus, fitting a mixture model with three components to such images could potentially model an unexisting negative activation.
Again GIM (blue) provides very sparse positive activation when the activation is strong (first row) but it is also the best estimating the absent activatio. On the other hand, bGGM (red) solutions are too dense, modelling non-existing activations specially at low SNR or sparse activations. GGM (pink) shows again the highest variance of all 4 models but it provides a good performance at high SNR even at very sparse activations (bottom rows, right sub-figures). GGM can severely overestimate activations. The bGIM (green) algorithm slightly overestimates the extremely sparse activations (rows 2 and 3) and provides good solutions for the most realistic activation density. Again, bGIM proves to be very robust.

\subsection{Resting State fMRI data}
\label{val_rfMRI}
\begin{figure}[!h]
\centering
\includegraphics[width=1\textwidth]{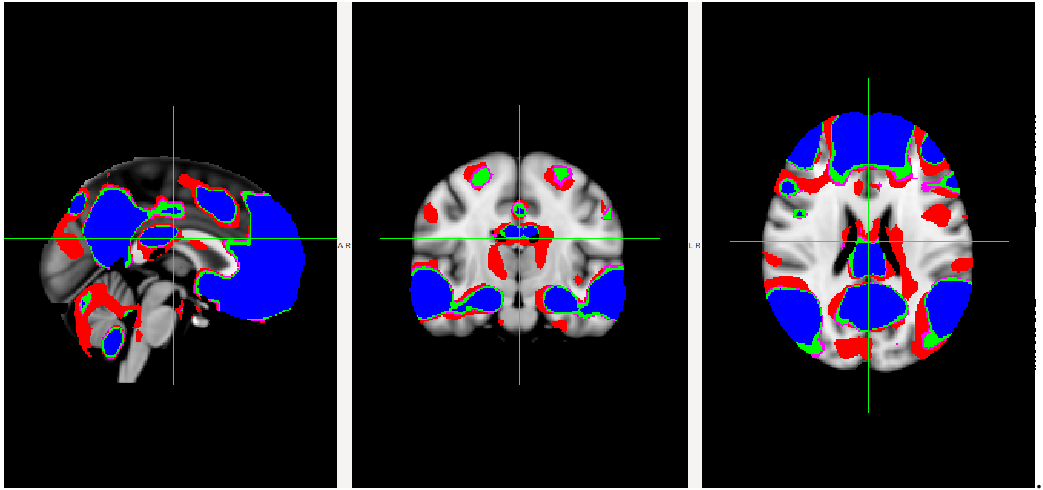}.
\caption{\footnotesize Example of activation maps provided by each algorithm. Color coded are as before.} 
\label{DMN}
\end{figure}

In this section we compare the four considered models when applying them to the 4400 statistical spatial maps derived from the resting-state fMRI data as described in section \ref{rfMRI}. Each image was masked to remove zero valued voxels and then standardized to zero mean and unit variance. We consider as active those voxels with a probability of activation bigger than a given threshold of p= 0.5. In Figure \ref{DMN}, we present the activation maps obtained by each of the four algorithms when evaluating a pseudo-random spatial map from the 1100 images at 1mm. Color coded are as before.
The bGGM model (red) provides the most dense solution, followed by GGM (pink) and bGIM (green). The sparsest solution is given by GIM (blue).
While bGGM provides a much denser solution than the other models, the difference between GGM and bGIM is moderate in this example. The solution provided by GIM is much sparser and it could omit interesting information as can be observed in the middle panel of figure \ref{DMN}; note that the symmetric superior activation reflected by all other models is neglected by GIM.  

\begin{figure}[!h]
\centering
\includegraphics[width=1\textwidth]{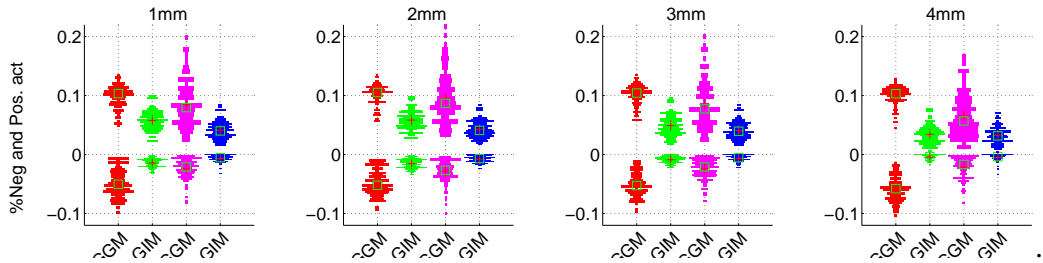}.
\caption{\footnotesize Each subplot shows violin plots of percentage of active voxels.The first row shows results for the positive component and the second row for the negative component.} 
\label{perc}
\end{figure}

To summarize the results obtained in the 4400 maps, in Figure \ref{perc} we show violin plots on the percentage of active voxels obtained by each algorithm (x-axis and color coded) at four different image resolutions as showed on each subfigure title. The proportion of negative active voxels is presented as a negative number.
All models agree in having more positive than negative activation. Independently of the image resolution, the GIM model provides the sparsest images followed by the bGIM; the most dense solutions are given by bGGM. The high variance in the GGM estimations shows that GGM probably overestimated the activation maps.

 %This is  followed by vGGM.  
%The model providing the most dense thresholded maps,vGGM, seems to clearly overestimate the activation. The other models follow the order of sparsity expected but clearly is difficult to mention which model provides better results.

\begin{figure}[!h]
\centering
\includegraphics[width=1\textwidth]{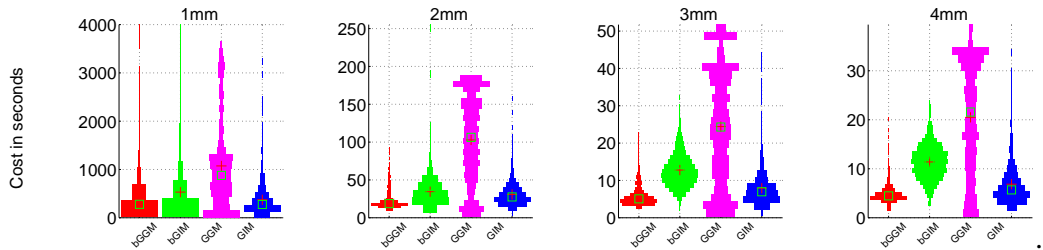}.
\caption{\footnotesize Each subplot shows violin plots of the algorithmic computational costs in seconds for a given image resolution.} 
\label{cost}
\end{figure}

Another important factor to keep on mind is the computational cost of each algorithms.
In figure \ref{cost} we show violin plots of the computational cost (in second) taken by each algorithm. From left to right we show the statistics obtained on the 1100 maps obtained at 1, 2 3, 4 mm MNI space respectively. 
We observe that bGGM is always the fastest followed by GIM and bGIM. The GGM is clearly the most computationally demanding with a cost distribution showing high variance; the cost is specially large for high image resolutions (left subfigure). The bGIM cost distribution is compact and its cost remains acceptable even for high resolution images.  

\section{Discussion}
\label{Discussion}
In this paper we reviewed the state-of-the-art algorithms for learning the parameters of mixture models containg Gamma (GGM) and inverse-Gamma components (GIM), and we introduced novel analytical variational Bayes learning procedures for these mixture models, denoted as bGGM and bGIM respectively. The updates for most model parameters are obtained using standard variational Bayes techniques; for the most involved ones we used Laplace approximations and Taylor expansions to compute the required expectations. We validated the performance of the algorithms in different simulated scenarios and extensive rfMRI data. As is usually done on rfMRI data, we fitted mixture models with three components (for both real and synthetic data). 

We observed that, in general, GIM provides the sparsest solutions, followed by bGIM, GGM; bGGM provides too dense solutions. The GGM solutions showed the highest variance of the four models and overestimated activations with respect to other models in the context of rfMRI data. GIM generally underestimates activations and the bGIM model provides an interesting intermediate solution. Evaluating the models using paired t-tests, bGIM turned out to be the best model in most cases.
When considering the computational cost we observed that bGGM is the fastest model closely followed by GIM and bGIM. All models enjoy significant computational advantages with respect to the previous state-of-the-art GGM, the difference becoming dramatic for high image resolutions. 

Put together, the bGIM model is an excellent candidate to replace GGM in many neuroimaging tasks. 
The presented variational methodology also allows the inclusion of Gamma or inverse-Gamma components in more complex inference problems,
for example extending VB mixture models for image segmentation \cite{vbmm1,vbmm2,vbmm3} to mixtures containing non-Gaussian components. In particular, it 
can be used to extend the work of \cite{WoolSPMM} to substitute the costly numerical optimization procedure for the Gamma parameters estimation. 
%Furthermore, the model can also containg inverse-Gamma components. 
Another important use of the presented models is in the context of variational ICA decompositions with a Gauss/Gamma or Gauss/inverse-Gamma source model. This assumption on the source model can enhance the sensitivity of the method by placing the source model assumption inside the learning procedure instead of as a post-hoc process.

\bibliographystyle{ieee}
\bibliography{bayesGammaMM.bib}

\newpage
\section{Appendices}
\appendix

\section{Variational mixture models}
Here we continue with the notation and the problem described in sections \ref{model_sec} and \ref{model_sec2}.
The joint probability density function is given by  
\begin{align*}
p(\boldsymbol{x},\boldsymbol{Z},\mu_1,\tau_1,\boldsymbol{s},\boldsymbol{r})=
\end{align*} 
\small
\begin{align}
\label{jointpdf}
=p(\boldsymbol{x}|\boldsymbol{Z},\mu_1,\tau_1,\boldsymbol{s},\boldsymbol{r}) p(\boldsymbol{Z}|\boldsymbol{\pi}) p(\boldsymbol{\pi})p(\mu_1)p(\tau_1) \prod_{k=2,3}p(s_k) p(r_k)
\end{align} 
\normalsize

The conditional distribution over $\boldsymbol{Z}$ given the mixing coefficients $\boldsymbol{\pi}$ is
\begin{align}
\label{condZ}
p(\boldsymbol{Z}|\boldsymbol{\pi})=\prod_{n=1}^{N} \prod_{k=1}^{3}  \pi_{k}^{z_{nk}}.
\end{align} 
The conditional distribution of the observations given the latent variables and each component parameters is

\begin{align*}
p(\boldsymbol{x}|\boldsymbol{Z},\mu_1,\tau_1,\boldsymbol{s},\boldsymbol{r})= 
\end{align*} 
\begin{align}
\label{condZ}
=\prod_{n=1}^{N}   \left[ p_1 (x_n|\mu_1,\tau_1)^{z_{n1}}  \prod_{k=2}^{3} p_2(x_n|s_k,r_k)^{z_{nk}} \right].
\end{align} 

\normalsize

Now, we introduce the priors over the parameters $\boldsymbol{\pi},\mu_1,\tau_1,s_2,r_2,s_3,r_3$.
The prior over the mixing proportions is symmetric Dirichlet $(\lambda_k= \lambda_0 \forall k \in \{1,2,3 \})$,
\begin{align*}
p (\boldsymbol{\pi})=  \mathcal{D}(\boldsymbol{\pi}| \lambda_0) = C(\lambda_0) \prod_{k=1}^{3} \pi_{k}^{\lambda_0 -1}.
\end{align*}
We use a Gaussian prior for the mean $\mu_1$ of the Gaussian component, parametrized using mean $m_0$ and precision $\tau_0$,
\begin{align*}
p (\mu_1)=  \mathcal{N}(\mu_1,|m_0,\tau_0),
\end{align*}
and a Gamma prior, parametrized using shape $c^0$ and scale $b^0$, for the precision $\tau_1$ 
\begin{align*}
p (\tau_1)=  \mathcal{G}_2(\tau_1|c^0,b^0).
\end{align*}
For the non-Gaussian components of the mixture model (second and third components) we use a Gamma prior over $\boldsymbol{r}=(r_2,r_3)$, parametrized using shape $d_0$ and rate $e_0$,   
\begin{align*}
p (\boldsymbol{r})=   \prod_{k=2}^{3} \mathcal{G}(r_k|d_0,e_0). 
\end{align*}
For the shape parameters $\boldsymbol{s}=(s_2,s_3)$, we use a prior of the form
\begin{align*}
p (\boldsymbol{s}) \propto  \prod_{k=2}^{3} p(s_k),
\end{align*}
where
\begin{align}
\label{GM_shape_prior}
p (s_k) \propto \frac{a_{0}^{s_k-1} r_{k}^{s_k c_{0}}}{\Gamma(s_k)^{b_{0}}} 
\end{align}
if component $k$ is Gamma distributed\footnote{$\Gamma$ denotes the Gamma function} and 
\begin{align}
\label{IG_shape_prior}
p (s_k) \propto \frac{a_{0}^{-s_k-1} r_{k}^{s_k c_{0}}}{\Gamma(s_k)^{b_{0}}} 
\end{align}
if component $k$ is inverse-Gamma distributed.

These functionals depend on the rates $\boldsymbol{r}$ and on three hyper parameters $(a_0,b_0,c_0)$. Equation (\ref{GM_shape_prior}) is an unnormailized conjugate prior for the shape of the Gamma distribution \cite{Fink97acompendium} and Equation (\ref{IG_shape_prior}) an unnormalized conjugate prior for the shape parameter of an inverse-Gamma distribution \cite{LleBe_arxiv_2016_1}.%(proof of conjugacy is straightforward). % by multiplying Inverse-Gamma likelihood by the presented functional and rearranging terms.

\subsection{Variational updates}
\label{VB}
We consider a variational distribution that factorizes between latent variables and parameters as
\begin{align*}
q(\boldsymbol{Z},\boldsymbol{\pi},\mu_1,\tau_1,\boldsymbol{s}, \boldsymbol{r})=q(\boldsymbol{Z}) q(\boldsymbol{\pi},\mu_1,\tau_1,\boldsymbol{s}, \boldsymbol{r}).
\end{align*}

\subsubsection{Latent variables}
Given a data vector of observations $\boldsymbol{x}=\{x_1,\ldots,x_N \}$, $x_i \in \mathbb{R}$ and
using standard VB results, we have that 
\footnotesize

\begin{align*}
\log q^{*}(\boldsymbol{Z}) = \langle \log p(\boldsymbol{x}, \boldsymbol{Z},\boldsymbol{\pi},\mu_1,\tau_1,\boldsymbol{s}, \boldsymbol{r}) \rangle_{\boldsymbol{\pi},\mu_1,\tau_1,\boldsymbol{s}, \boldsymbol{r}} + \mathrm{const}.
\end{align*}
\normalsize
Considering Equation (\ref{jointpdf}) and keeping only terms that depend on $\boldsymbol{Z}$, we have that
\begin{align*}
\log q^{*}(\boldsymbol{Z}) = 
\end{align*}
\small
\begin{align*}
= \langle \log p( \boldsymbol{Z}|\boldsymbol{\pi}) \rangle_{\boldsymbol{\pi}} +  \langle[ \log p( \boldsymbol{x}|\boldsymbol{Z},\mu_1,\tau_1, \boldsymbol{r},\boldsymbol{s} ) \rangle_{\mu_1,\tau_1,\boldsymbol{r},\boldsymbol{s} } +  \mathrm{const}.
\end{align*}
\normalsize

Substituting conditionals and absorbing terms that are independent from $\boldsymbol{Z}$ into the constant term, we obtain
\begin{align}
\log q^{*}(\boldsymbol{Z}) = \sum_{n=1}^{N} \sum_{k=1}^{3} z_{nk} \log \rho_{nk} +  \mathrm{const},
\label{qZ}
\end{align}
where 
\begin{align*}
\log \rho_{n1} = \langle\log \pi_1 \rangle + \frac{1}{2} \langle \log \tau_1 \rangle- \frac{1}{2} \log(2\pi) +%- \frac{1}{2}  \langle(x_n-\mu_1)^2\rangle_{\mu_1} \langle \tau_1 \rangle.
\end{align*}
\begin{align*}
- \frac{1}{2}  \langle(x_n-\mu_1)^2\rangle_{\mu_1} \langle \tau_1 \rangle.
\end{align*}

For the Gamma components we have
 \begin{align*}
\log \rho_{nk} = \langle \log \pi_k \rangle + (\langle s_k \rangle -1)\log(x_n) + \langle s_k \rangle \langle \log r_k \rangle +%- \langle \log \Gamma(s_k) \rangle - \langle r_k \rangle x_n.
\end{align*} 
\begin{align*}
 - \langle \log \Gamma(s_k) \rangle - \langle r_k \rangle x_n.
\end{align*} 
For inverse-Gamma components we have
\begin{align*}
\log \rho_{nk} = \langle \log \pi_k \rangle - (\langle s_k \rangle +1)\log(x_n) + \langle s_k \rangle \langle \log r_k \rangle +%- \langle \log \Gamma(s_k) \rangle - \frac{\langle r_k \rangle}{ x_n}. 
\end{align*} 
\begin{align*}
 - \langle \log \Gamma(s_k) \rangle - \frac{\langle r_k \rangle}{ x_n}. 
\end{align*}

Due to the positive support of the Gamma/inverse-Gamma distributions and the negative support of negative Gamma/inverse-Gamma distributions, we define $\log \rho_{nk} = - \infty $ if $x_n < 0$ and component $k$ is positive or, if $x_n > 0$ and component $k$ is negative.

Exponentiating both sides of (\ref{qZ}) we have
\begin{align*}
q^{*}(\boldsymbol{Z}) \propto \prod_{n=1}^{N} \prod_{k=1}^{3}  \rho_{nk}^{z_{nk}},
\end{align*}
so 
\begin{align*}
q^{*}(\boldsymbol{Z}) = \prod_{n=1}^{N} \prod_{k=1}^{3}  \gamma_{nk}^{z_{nk}},
\end{align*}
where 
\begin{align*}
\gamma_{nk} = \frac{\rho_{nk}}{\sum_{j=1}^{3}\rho_{nj}}.
\end{align*}
%so the quantities $\gamma_{nk}$ play the role of the responsibilities \cite{bishop}.

\subsubsection{Model parameters}
Turning to the functional $q(\boldsymbol{\pi},\mu_1,\tau_1,\boldsymbol{s}, \boldsymbol{r})$,
we now derive the VB updates for the parameters $w\in\{\boldsymbol{\pi}, \mu_1,\tau_1, \boldsymbol{r},\boldsymbol{s} \}$.

First, we define  
\begin{align*}
N_k = \sum_{n=1}^{N} \gamma_{nk},
\end{align*}
\begin{align*}
\bar{\boldsymbol{x}}_k = \sum_{n=1}^{N} \gamma_{nk} x_n.
\end{align*}

Taking the expectations over $\boldsymbol{Z}$ we have that
\begin{align*}
\log q^*(\boldsymbol{\pi},\mu_1,\tau_1,\boldsymbol{s}, \boldsymbol{r}) =  \langle \log p(\boldsymbol{x},\boldsymbol{\theta}) \rangle_{\boldsymbol{Z}} + \mathrm{const} =
\end{align*}
\begin{align*}
 =  \sum_{n=1}^{N}  \left[ \langle z_{n1} \rangle \log p_1(x_n|\mu_1,\tau_1) + \sum_{k=2}^3 \langle z_{nk} \rangle \log p_k (x_n|s_k,r_k) \right] + % + \langle z_{n3} \rangle \log p_3 (x_n|s_3,r_3) 
\end{align*}
\begin{align*}
+\log p(\boldsymbol{\pi}) +  \langle \log p(\boldsymbol{Z}|\boldsymbol{\pi}) \rangle_{\boldsymbol{Z}} + % \log p(\mu_1) + \log p(\tau_1) + \sum_{k=2}^3 \log p(s_k,r_k). 
\end{align*}
\begin{align}
 +  \log p(\mu_1) + \log p(\tau_1) + \sum_{k=2}^3 \log p(s_k,r_k).
\label{q*} 
\end{align}

This expression is used to derive the parameter updates in the following subsections. In particular, for a given parameter $w\in\{\boldsymbol{\pi}, \mu_1,\tau_1, \boldsymbol{r},\boldsymbol{s} \}$, we identify terms in (\ref{q*}) that depend on $w$ to get an expression for $\log q^*(w)$. Exponentiating and regrouping terms lead us to the rest of the updates.  

For $\boldsymbol{\pi}$, we have

%Identifying terms in (\ref{q*}) that depend on $\boldsymbol{\pi}$ we get an expression for $\log q^*(\boldsymbol{\pi})$. Exponentiating and gathering terms we obtain
\begin{align*}
q^*(\boldsymbol{\pi}) = \mathcal{D}(\boldsymbol{\pi}|\hat{\boldsymbol{\lambda}}),
 \end{align*}
\begin{align*}
\hat{\lambda}_k = \lambda_0 + N_k.
 \end{align*}

For  $\mu_1$, we have 

%Identifying terms in (\ref{q*}) that depend on $\mu_1$ we get an expression for $\log q^*(\mu_1)$. Exponentiating and gathering terms we obtain
\begin{align*}
q^*(\mu_1) = \mathcal{N}(\mu|\hat{m},\hat{\tau}),
\end{align*}
\begin{align*}
\hat{m}=
\frac{1}{\hat{\tau}} \left( \tau_0 m_0 + \langle \tau_1 \rangle \bar{\boldsymbol{x}}_1 \right),
\end{align*}
\begin{align*}
\hat{\tau}= 
\tau_0 + \langle\tau_1 \rangle N_1.
\end{align*}

For $\tau_1$, we have
%Identifying terms in (\ref{q*}) that depend on $\tau_1$ we get an expression for $\log q^*(\tau_1)$. Exponentiating and gathering terms we obtain
\begin{align*}
%\label{C}
q^*(\tau_1)   =  \mathcal{G}(\tau_1 | \hat{c},  \hat{b}),
\end{align*}
\begin{align*}
\hat{b}= \left[ \frac{1}{b^0} + \frac{1}{2} \sum_{n=1}^{N} \gamma_{n1} ( x_n^2 + \langle\mu_1^2 \rangle - 2 x_n \langle\mu_1 \rangle)  \right]^{-1},
\end{align*}
\begin{align*}
\hat{c}= 
c^0 + \frac{1}{2} N_1.
\end{align*}

For $\boldsymbol{r}$, we have
%Identifying terms in (\ref{q*}) that depend on $\boldsymbol{r}$ we get an expression for $\log q^*(\boldsymbol{r})$. Exponentiating and gathering terms we obtain
\begin{align*}
q^*(\boldsymbol{r})  = \prod_{k=2}^{3} \mathcal{G}_2(r_k | \hat{d}_k,\hat{e}_k),
\end{align*}
\begin{align*}
\hat{d}_k=
 d_0 +  \langle s_k \rangle N_k,
\end{align*}
\begin{align*}
\hat{e}_k=
e_0 + \bar{\boldsymbol{x}}_k.
\end{align*}

For $\boldsymbol{s}$, we have, for Gamma components
\begin{align*}
q^*(s_k) \propto  \frac{\hat{a}_k^{s_k -1} r_k^{s_k\hat{c}_k}}{\Gamma(s_k)^{\hat{b}_k}},
\end{align*}
and for inverse-Gamma components
\begin{align*}
q^*(s_k) \propto  \frac{\hat{a}_k^{-s_k -1} r_k^{s_k\hat{c}_k}}{\Gamma(s_k)^{\hat{b}_k}}. 
\end{align*}
In both cases we have
\begin{align*}
\hat{a}_k = a_0 \prod_{n=1}^{N} x_n^{\gamma_{nk}},
\end{align*}
\begin{align*}
\hat{b}_k = %b_0+ \sum_{n=1}^N \gamma_{nk} = 
b_0+ N_k,
\end{align*}
\begin{align*}
\hat{c}_k = %c_0+\sum_{n=1}^N \gamma_{nK}= 
c_0+ N_k.
\end{align*}

\subsection{Computing the expectations}
Using standard results for a Dirichlet distribution we have that for $k \in \{1,2,3\}$ the posterior required expectations over $\pi_k$ are given by
\begin{align*}
\langle\pi_k \rangle= \frac{\hat{\lambda}_k}{\sum_{m=1}^3 \hat{\lambda}_m},
\end{align*}
\begin{align*}
\langle\log \pi_k \rangle = \Psi(\hat{\lambda}_k) - \Psi(\sum_{m=1}^3 \hat{\lambda}_m).
\end{align*}

Using standard results for a Gaussian distribution we have that the required posterior expectations over $\mu_1$ are
\begin{align*}
\langle\mu_1 \rangle = \hat{m},
\end{align*}
\begin{align*}
\langle\mu_1^2 \rangle = \hat{m}^2 + \frac{1}{\hat{\tau}}.
\end{align*}

Using standard results for a Gamma distribution we have that the required posterior expectations over $\tau_1$ are
\begin{align*}
\langle\tau_1 \rangle = \hat{b} \hat{c},
\end{align*}
\begin{align*}
\langle\tau_1^2 \rangle= \hat{b} \hat{c} (1+\hat{c}),
\end{align*}
\begin{align*}
\langle\log \tau_1 \rangle = \Psi(\hat{c}) + \log \hat{b},
\end{align*}
and considering a Gamma distribution parametrized using shape and rate we obtain the required posterior expectations over $r_k$ for $k \in \{2,3 \}$
\begin{align*}
\langle r_k \rangle = \frac{\hat{d}_k}{\hat{e}_{k}},
\end{align*}
\begin{align*}
\langle\log r_k \rangle =\Psi(d)-\log e. 
\end{align*}
We compute the required expectations over $\boldsymbol{s}$ using the Laplace approximation. Consider the prior on the Gamma shape with the form of equation (\ref{GM_shape_prior}),
\begin{align*}
p_{G}(s|a, b, c, r)   \propto  \frac{a^{s-1} r^{s c}}{\Gamma(s)^{b}} 
\end{align*}
and the prior on the inverse-Gamma shape with the form of equation (\ref{IG_shape_prior}),
\begin{align*}
p_{IG}(s|a, b, c, r)   \propto  \frac{a^{-s-1} r^{s c}}{\Gamma(s)^{b}} 
\end{align*}

Making use of the chain rule we have
\begin{align*}
\frac{d \log p(s|a, b, c, r)}{ds}=  \frac{d \log p(s|a, b, c, r)}{d p(s|a,b,c,r)} \frac{d p(s|a,b,c,r)}{d s},
\end{align*}
and, since,
\begin{align}
\frac{d p_{G}(s|a, b, c, r)}{ds}=  p_{G}(s) [  \log a + c \log r - b \Psi(s)  ],
\label{d1fancy}
\end{align}
and

\begin{align}
\label{d1fancy2}
\frac{d p_{IG}(s|a, b, c, r)}{ds}=  p_{IG}(s) [ - \log a + c \log r - b \Psi(s)  ],
\end{align}

we have that
\begin{align*}
\frac{d \log p_{G}(s|a, b, c, r)}{ds}=   \log a + c \log r - b \Psi(s)
\end{align*}
and 
\begin{align*}
\frac{d \log p_{IG}(s|a, b, c, r)}{ds}=   -\log a + c \log r - b \Psi(s).
\end{align*}
Further, both second derivatives are equal for both cases
\begin{align*}
\frac{d^2 \log p_{G}(s|a, b, c, r)}{d^2s}= \frac{d^2 \log p_{IG}(s|a, b, c, r)}{d^2s} =- b \Psi_1(s),
\end{align*}
where $\Psi_{1}(s) =\frac{d \Psi(s)}{ds}$.
Therefore %the Laplace approximation to $p(s|a,b,c,r)$ is a Gaussian with mean $\mu$ and precision $b\Psi_1(k)$
\begin{align*}
p_{G}(s|a,b,c,r) \approx \mathcal{N}(s|\mu_{G},b\Psi_1(\mu))
\end{align*}
and 
\begin{align*}
p_{IG}(s|a,b,c,r) \approx \mathcal{N}(s|\mu_{IG},b\Psi_1(\mu)),
\end{align*}
where 
\begin{align*}
\mu_{G}= \Psi^{-1}\Big(\frac{\log a + c \log r}{b}\Big)
\end{align*}
is a zero of (\ref{d1fancy}) and 
\begin{align*}
\mu_{IG}= \Psi^{-1}\Big(\frac{-\log a + c \log r}{b}\Big)
\end{align*}
is a zero of (\ref{d1fancy2}). 

Using these approximations we have that the first required expectation is approximated in the case of the Gamma by
\begin{align*}
\langle s_k \rangle \approx \Psi^{-1} \Big( \frac{\log a_k + c_k \log r_k}{b_k}\Big)
\end{align*}
and, in the one of the inverse-Gamma, by 
\begin{align*}
\langle s_k \rangle \approx \Psi^{-1}\Big(\frac{-\log a_k + c_k \log r_k}{b_k}\Big).
\end{align*}

The other required expectation is $\mathbb{E}[\log(\Gamma(s_k)]$. We use Taylor expansion 
to obtain 
\begin{align*}
\mathbb{E}[\log(\Gamma(s)]\approx \mathbb{E}[\log(\Gamma(\mu)] + \frac{1}{b} +   \frac{\Psi_2(\mu) \mu}{\Psi_1(\mu) b}. 
\end{align*}

\subsection{Hyper-parameters and initialization.}
\label{HP}
For the Gaussian component, we fixed the hyper-prior parameters values at $m_0=0$, $\tau_1 =1$, $c^0=0.01$ and $b^0=100$. This ensures that the mean is approximately centered at zero with a flat prior for the variance. For the Gamma (or inverse Gamma) components, we chose a prior distributions such that both mean and variance are set to 10. We then use the method of moments (see Appendix A) to estimate the prior distribution parameters ($s_0$ shape and $r_0$ rate/scale).  We set $d_0=r_0$ and $e_0=1$, so that the variance on $r_0$  has the same magnitude. For the hyper-priors on the shape parameter, $s_0$, we use the Laplace approximation (see Appendix A.2) to define a prior with the required expected value (and variance), resulting in 
\begin{align*}
b_0=c_0=\frac{1}{s_0 \Psi_1(s_0)}.
\end{align*}
For Gamma components, we have 
\begin{align*}
\log a_0=b_0 \Psi(s_0)-c_0 \log r_0.
\end{align*}
For inverse-Gamma components, we have  
\begin{align*}
\log a_0=-b_0 \Psi(s_0)+c_0 \log r_0.
 \end{align*} 
Finally the prior over the mixing proportions is fixed to $\lambda_0=5$.

The mixture model parameter initialization is performed using k-means \cite{MB}. The estimated means and variances are transformed into parameters for Gamma or inverse Gamma distributions for the required components using the method of moments (see Appendix B). These parameters are also used to estimate the density of each sample with respect to the non-Gaussian components required to estimate all initial $\gamma_{nk}$. 

\subsection{Convergence}
\label{CV}
The convergence of the algorithms are monitored using the negative free energy (NFE). The NFE for the proposed model is given by 
\footnotesize

\begin{align*}
F=   \langle \log p(\boldsymbol{x}, \boldsymbol{Z}|\boldsymbol{\pi},\mu_1,\tau_1,\boldsymbol{s}, \boldsymbol{r}) \rangle_{\boldsymbol{Z},\boldsymbol{\pi},\mu_1,\tau_1,\boldsymbol{s}, \boldsymbol{r}} + 
\end{align*}
\begin{align*}
+\mathcal{H}[q^*(\boldsymbol{Z})]-\mathcal{KL}[\boldsymbol{\pi}] - \mathcal{KL}[\mu_1] - \mathcal{KL}[\tau_1] -\mathcal{KL}[\boldsymbol{s}] -\mathcal{KL}[\boldsymbol{r}].  
\end{align*}
\normalsize
The joint-likelihood (averaged over the posteriors) and the entropy term are straightforward to obtain. 
The $\mathcal{KL}$-divergences between priors and posteriors can be found elsewhere \cite{choud}. 
The only cumbersome term is $\mathcal{KL}[\boldsymbol{s}]$ which does not have a known analytical solution.
We therefore approximate the $\mathcal{KL}$-divergence by the $\mathcal{KL}$-divergence between the Gaussian approximations obtained by the Laplace approximations to $p(\boldsymbol{s})$ (see Appendix A2).

\section{Method of Moments}
Given a data vector of observations $\boldsymbol{x}=\{x_1,\ldots,x_N \}$, $x_i \in \mathbb{R}$, and defining $\mu$ as the mean of $\boldsymbol{x}$ and $v$ as its variance, the method of moments parameters estimation for the Gamma distribution reads
\begin{align*}
s \approx \frac{\mu^2}{v},  \hspace{35pt}  \frac{1}{r}\approx\frac{v}{\mu}.
\end{align*} 
where s is the shape parameter and r is the rate parameter, and  
\begin{align*}
s \approx  \frac{\mu^2}{v}+2,  \hspace{35pt} r \approx \mu (\frac{\mu^2}{v}+1). 
\end{align*}
for the inverse-Gamma distribution where s is the shape parameter and r is the scale parameter.

\section{State of the art mixture models}
Algorithm 3 summarizes the approximated maximum likelihood algorithm presented in \cite{poster,fsl} for learning a Gaussian/Gamma mixture model (GGM).
\begin{algorithm}
\caption*{Algorithm 3: ML Gauss Gamma mixture model (GGM)} \label{alg:GGM}
\begin{algorithmic}[1]
\REQUIRE  \small Data: $\textbf{x}=\{x_1,\ldots,x_N \}, x_n \in \mathbb{R}$; \newline
Parametrization: $p(x_n|\Theta, \Pi)=  \sum_{k=1}^{K}\pi_{k} p_k(x_n|\theta_k)$ \newline
\scriptsize $p_1(x_n|\Theta_1)= \mathcal{N}(x_n|\mu_1,v_1)$, $p_2(x_n|\Theta_2)= IG(x_n|s_2,r_2)$, $p_3(x_n|\Theta_K) = IG^{-}(x_n|s_3,r_3)$.
\STATE \small Initialization parameter values: $\Theta = \{ \mu_1,v_1, s_2,r_2, s_3, r_3\},\Pi=\{\pi_1,\pi_2,\pi_3 \}$
\REPEAT
\FOR{$n \in \{1,\ldots,N \}$}
\FOR{$k \in \{1,\ldots,3 \}$}
\STATE $\gamma_{k}(x_n) = \frac{\pi_{k} p_k(x_n|\Theta_m) }{\sum_{j=1}^{3} \pi_{j} p_j(x_n|\Theta_j)}.$
\ENDFOR
\ENDFOR
\FOR{$k \in \{1,\ldots,3 \}$}
\STATE $\mu_k= \frac{1}{N_k} \sum_{n=1}^{N} \gamma_{k}(x_n) x_n$
\STATE $v_k= \frac{1}{N_k} \sum_{n=1}^{N} \gamma_{k}(x_n)(x_n- \mu_k)^2$
\IF{$k \in \{2,3 \}$ }
\STATE ${\alpha_k} = $%\frac{\mu_k^2}{v_k}+2,\hspace{5pt} {\beta_k} = \mu_k (\frac{\mu_k^2}{v_k}+1)$
\ENDIF
\STATE $N_k=\sum_{n=1}^{N}\gamma_{k}(x_n)$
\STATE $\pi_{k} = \frac{N_k}{\sum_{j=1}^{3} N_j}$
\ENDFOR
\UNTIL {convergence}
\RETURN $\Theta$, $\Pi$.
\normalsize
\end{algorithmic}
%\caption{Gauss Gamma mixture model (GGM)}
\end{algorithm}

Algorithm 4 summarizes the algorithm presented in \cite{posterohbm2015} for learning a Gaussian/inverse-Gamma mixture model (GIM).
\begin{algorithm}
\caption*{Algorithm 4: ML Gauss inverse-Gamma mixture model (GIM)} \label{alg:GIM}
\begin{algorithmic}[1]
\REQUIRE  \small Data: $\textbf{x}=\{x_1,\ldots,x_N \}, x_n \in \mathbb{R}$; \newline
Parametrization: $p(x_n|\Theta, \Pi)=  \sum_{k=1}^{K}\pi_{k} p_k(x_n|\theta_k)$ \newline
\scriptsize $p_1(x_n|\Theta_1)= \mathcal{N}(x_n|\mu_1,v_1)$, $p_2(x_n|\Theta_2)= IG(x_n|s_2,r_2)$, $p_3(x_n|\Theta_K) = IG^{-}(x_n|s_3,r_3)$.
\STATE \small Initialization parameter values: $\Theta = \{ \mu_1,v_1, s_2,r_2, s_3, r_3\},\Pi=\{\pi_1,\pi_2,\pi_3 \}$
\REPEAT
\FOR{$n \in \{1,\ldots,N \}$}
\FOR{$k \in \{1,\ldots,3 \}$}
\STATE $\gamma_{k}(x_n) = \frac{\pi_{k} p_k(x_n|\Theta_m) }{\sum_{j=1}^{3} \pi_{j} p_j(x_n|\Theta_j)}.$
\ENDFOR
\ENDFOR
\FOR{$k \in \{1,\ldots,3 \}$}
\STATE $\mu_k= \frac{1}{N_k} \sum_{n=1}^{N} \gamma_{k}(x_n) x_n$
\STATE $v_k= \frac{1}{N_k} \sum_{n=1}^{N} \gamma_{k}(x_n)(x_n- \mu_k)^2$
\IF{$k \in \{2,3 \}$ }
\STATE ${\alpha_k} = \frac{\mu_k^2}{v_k}+2,\hspace{5pt} {\beta_k} = \mu_k (\frac{\mu_k^2}{v_k}+1)$
\ENDIF
\STATE $N_k=\sum_{n=1}^{N}\gamma_{k}(x_n)$
\STATE $\pi_{k} = \frac{N_k}{\sum_{j=1}^{3} N_j}$
\ENDFOR
\UNTIL {convergence}
\RETURN $\Theta$, $\Pi$.
\normalsize
\end{algorithmic}
%\caption{Gauss inverse-Gamma mixture model (GIM)}
\end{algorithm}

\end{document}